\documentclass[sigconf]{acmart}
\AtBeginDocument{%
  }

\copyrightyear{2025}
\acmYear{2025}
\setcopyright{cc}
\setcctype{by}
\acmConference[MM '25]{Proceedings of the 33rd ACM International Conference on Multimedia}{October 27--31, 2025}{Dublin, Ireland}
\acmBooktitle{Proceedings of the 33rd ACM International Conference on Multimedia (MM '25), October 27--31, 2025, Dublin, Ireland}
\acmDOI{10.1145/3746027.3755118}
\acmISBN{979-8-4007-2035-2/2025/10}

\usepackage{graphicx}
\usepackage[linesnumbered,ruled]{algorithm2e}
\usepackage{multirow}
\usepackage{pifont}
\usepackage{colortbl}
\usepackage{xspace}

\newcommand{\eg}{\textit{e.g.}\xspace}
\newcommand{\ie}{\textit{i.e.}\xspace}

\newcommand{\etc}{\textit{etc.}\xspace}
\usepackage{enumitem}
\AtEndPreamble{
    \usepackage[capitalize]{cleveref}
    \crefname{section}{Sec.}{Secs.}
    \Crefname{section}{Section}{Sections}
    \crefname{table}{Tab.}{Tabs.}
    \Crefname{table}{Table}{Tables}
}
\usepackage{amsmath}
\usepackage{booktabs}
\usepackage{subcaption}

\begin{document}

\title{DHCP: \underline{D}etecting \underline{H}allucinations by \underline{C}ross-modal Attention \underline{P}attern in Large Vision-Language Models}

\author{Yudong Zhang}
\orcid{0009-0009-6049-603X}

\affiliation{%
  \institution{Tsinghua University, Tencent}
  \city{Beijing}
  \country{China}
}
\email{zhangyd16@mails.tsinghua.edu.cn}

\author{Ruobing Xie}
\orcid{0000-0003-3170-5647}
\authornote{Corresponding authors.}
\affiliation{%
  \institution{Tencent}
  \city{Beijing}
  \country{China}}
\email{xrbsnowing@163.com}

\author{Xingwu Sun}
\orcid{0009-0008-3222-0901}
\affiliation{%
 \institution{Tencent, University of Macau}
 \city{Beijing}
 \country{China}}
\email{sunxingwu01@gmail.com}

\author{Yiqing Huang}
\orcid{0000-0002-2143-3329}
\affiliation{%
 \institution{Tencent}
 \city{Beijing}
 \country{China}}
\email{huang-yq17@tsinghua.org.cn}

\author{Jiansheng Chen}
\orcid{0000-0002-2040-7938}
\authornotemark[1]
\affiliation{%
  \institution{University of Science and Technology Beijing}
  \city{Beijing}
  \country{China}
}
\email{jschen@ustb.edu.cn}

\author{Zhanhui Kang}
\orcid{0009-0006-5151-4222}
\affiliation{%
 \institution{Tencent}
 \city{Shenzhen, Guangdong}
 \country{China}}
\email{kegokang@tencent.com}

\author{Di Wang}
\orcid{0009-0003-6713-1638}
\affiliation{%
 \institution{Tencent}
 \city{Beijing}
 \country{China}}
\email{diwang@tencent.com}

\author{Yu Wang}
\orcid{0000-0001-6108-5157}
\authornotemark[1]
\affiliation{%
  \institution{Tsinghua University}
  \city{Beijing}
  \country{China}}
\email{yu-wang@mail.tsinghua.edu.cn}

\renewcommand{\shortauthors}{Zhang et al.}

\begin{abstract}
Large vision-language models (LVLMs) have demonstrated exceptional performance on complex multimodal tasks. However, they continue to suffer from significant hallucination issues, including object, attribute, and relational hallucinations. To accurately detect these hallucinations, we investigated the variations in cross-modal attention patterns between hallucination and non-hallucination states. Leveraging these distinctions, we developed a lightweight detector capable of identifying hallucinations. Our proposed method, Detecting Hallucinations by Cross-modal Attention Patterns (DHCP), is straightforward and does not require additional LVLM training or extra LVLM inference steps. Experimental results show that DHCP achieves remarkable performance in hallucination detection. By offering novel insights into the identification and analysis of hallucinations in LVLMs, DHCP contributes to advancing the reliability and trustworthiness of these models. The code is available at \url{https://github.com/btzyd/DHCP}.
\end{abstract}




\maketitle

\section{Introduction}
\label{sec:intro}

Recent breakthroughs in large language models (LLMs) \cite{zhang2022opt, chung2024scaling}, such as Vicuna \cite{chiang2023vicuna}, Qwen \cite{yang2024qwen2}, and LLaMA \cite{touvron2023llama}, have led to the emergence of a prominent class of large vision-language models (LVLMs). Through the incorporation of aligned visual tokens as part of their input, these models have demonstrated remarkable capabilities in handling vision-language tasks with considerable effectiveness \cite{li2023blip, zhu2023minigpt}. Prominent examples within this category include InstructBLIP \cite{dai2024instructblip}, Qwen-VL \cite{bai2023qwen, bai2025qwen2}, Intern-VL \cite{chen2024internvl} and LLaVA \cite{liu2024visual}. Notwithstanding these impressive accomplishments, a critical challenge persists: the issue of \textbf{\emph{hallucinations}}.

\begin{figure}[!t]
  \centering
  \begin{subfigure}{\linewidth}
        \includegraphics[width=\linewidth]{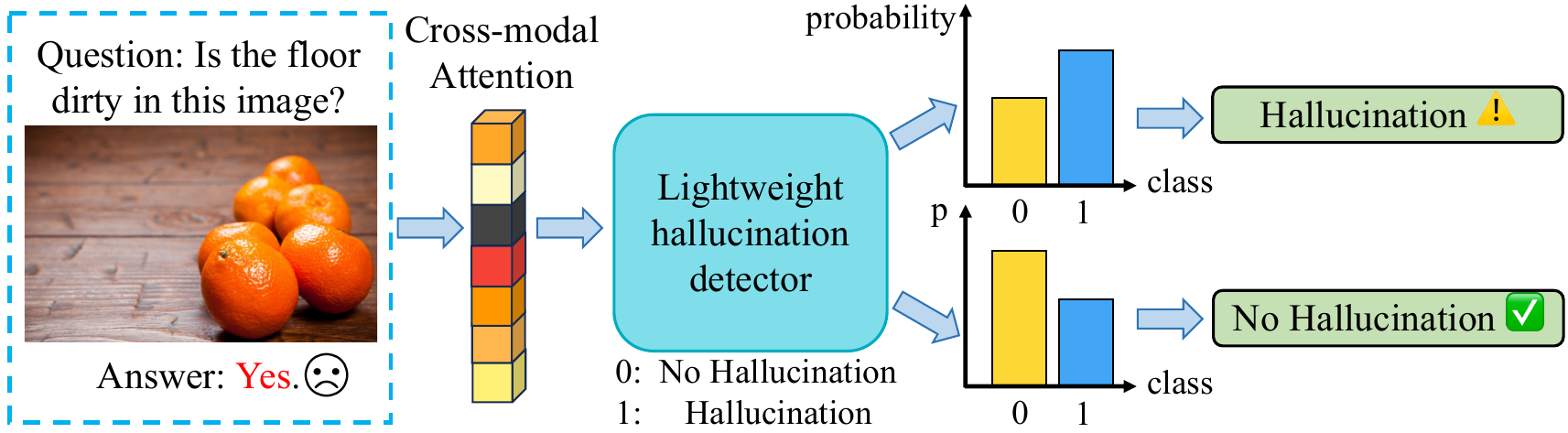}
        \caption{DHCP effectively identifies various hallucinations by employing a lightweight detector to monitor the cross-modal attention of LVLMs.}
        \label{fig:fig1c}
    \end{subfigure}
    \hfill
  \begin{subfigure}{\linewidth}
        \includegraphics[width=\linewidth]{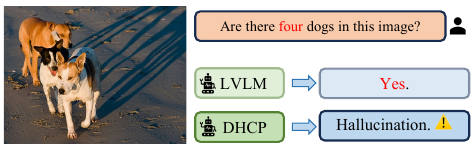}
        \caption{DHCP detects hallucinations in discriminative tasks. This is an example of number hallucination within attribute hallucination.}
        \label{fig:fig1a}
    \end{subfigure}
    \hfill
    \begin{subfigure}{\linewidth}
        \includegraphics[width=\linewidth]{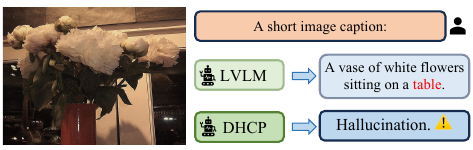}
        \caption{DHCP detects hallucinations in generative tasks. There is no table in the image, and the model depicts it as such.}
        \label{fig:fig1b}
    \end{subfigure}
    \caption{The schematic diagram and application of DHCP.}
    \label{fig:fig_1}
\end{figure}

Current approaches for assessing hallucinations in LVLMs can be broadly categorized into two main types: discriminative and generative methods. Discriminative methods, such as POPE \cite{li2023evaluating, lovenia2023negative, hu2023ciem}, involve framing objects, attributes, or relationships as yes/no questions, like ``Is there a/an {object}?'' The severity of hallucinations is then assessed based on the model's responses to these queries. On the other hand, generative methods \cite{gunjal2024detecting, liu2023mitigating, jing2023faithscore, wang2023evaluation, sun2023aligning, wang2023llm} assess hallucinations by analyzing task-specific generation performance.

As we explore the mechanisms behind hallucinations in LVLMs, one potential explanation lies in the model's misaligned focus on visual inputs. Specifically, the model may erroneously concentrate on certain image features, thereby perceiving objects or patterns that are not actually present. To analyze this phenomenon more systematically, we formally define \textbf{cross-modal attention} as \emph{the attention patterns of generated text tokens relative to visual tokens}. This definition enables us to rigorously examine how cross-modal interactions unfold within LVLMs. We hypothesize that when the model enters a hallucinatory state, its attention patterns toward visual tokens may differ significantly from those observed in non-hallucinating states. Such distinguishable attention signatures could provide critical insights for detecting model hallucinations effectively.

Our objective is to specifically detect hallucinations in LVLMs through the proposed DHCP method, as demonstrated in \cref{fig:fig1c}. Without employing DHCP, LVLMs tend to exhibit hallucinatory behaviors, such as producing erroneous outputs in discriminative tasks (\cref{fig:fig1a}) or generating content with discrepancies from input images in generative tasks (\cref{fig:fig1b}). By integrating DHCP during LVLM inference, the method monitors cross-modal attention patterns in real-time to identify whether the model is entering a hallucinatory state. Through this analytical process, DHCP triggers warnings for potentially hallucinatory outputs or entirely blocks such generations when necessary. Given the increasing popularity of test-time scaling \cite{muennighoff2025s1}, DHCP can also be integrated with TTS methods, such as generating multiple answers through multiple sampling to identify and eliminate responses detected as hallucinations by DHCP. Notably, DHCP provides an efficient and practical solution for hallucination detection without requiring additional training of the LVLMs or introducing extra inference steps of LVLMs, while demonstrating impressive robustness in identifying hallucinations. Additionally, the structure of the DHCP detector is straightforward, comprising a two-layer MLP that incorporates cross-modal attention of the LVLM generation process as input, resulting in an exceptionally low computational overhead during inference of hallucination detection.

We evaluated DHCP on several popular LVLMs, including InstructBLIP, Qwen2.5-VL, InterlVL-2.5 and LLaVA-v1.5. Our experiments demonstrate that DHCP exhibits robust performance in detecting hallucinations across both discriminative and generative tasks. Additionally, our analysis reveals that most false alarms generated by DHCP occur in cases where the model exhibits uncertainty in its responses, often nearing randomness. These instances warrant attention and necessitate further investigation to enhance the reliability and trustworthiness of LVLMs' outputs. Our findings also suggest that traditional evaluation metrics may underestimate DHCP's true effectiveness. This is because conventional metrics often overlook scenarios in which model uncertainty is misclassified as false positives. Furthermore, we explored DHCP's potential for categorizing the origins of hallucinations within LVLMs, aiming to uncover insights into the underlying causes of different types of hallucinatory outputs. 

Our main contributions can be summarized as follows: (1) We uncover significant differences in cross-modal attention patterns between hallucination and non-hallucination samples within LVLMs. (2) We propose DHCP, a novel, simple, effective, and efficient approach for distinguishing hallucination samples from the others by analyzing cross-modal attention patterns. DHCP eliminates the need for additional training or complex reasoning about LVLMs; instead, it employs a lightweight detector trained to monitor and evaluate the cross-modal attention of LVLMs. (3) Through comprehensive experiments across both discriminative and generative hallucination evaluation tasks, we demonstrate that DHCP consistently achieves strong performance in detecting hallucinations, showcasing its robustness and reliability.
We believe that DHCP could facilitate practical LVLM usages against hallucinations.

\section{Related Works}
\label{sec:ro}

\subsection{Large Vision-Language Model (LVLMs)}
LVLMs are typically composed of three core components: a visual encoder, a vision-language model, and a projector. The visual encoder \cite{radford2021learning, fang2023eva, zhang2025enhancing} processes input images to generate visual features. The projector then aligns these visual features with the input space of the LLM, enabling the LLM to interpret and process visual information effectively. Common implementations of the projector \cite{zhu2025connector, zhang2025security} include cross-attention mechanisms \cite{alayrac2022flamingo}, linear layers or multi-layer perceptrons (MLPs) \cite{liu2024visual, liu2023improved, chen2023minigpt}, adapters \cite{gao2023llama}, and Q-former architectures \cite{zhu2023minigpt, li2023blip, dai2024instructblip}. For the LLM component, several pre-trained models are available for selection, such as LLaMA \cite{touvron2023llama}, Vicuna \cite{chiang2023vicuna}, and Qwen \cite{yang2024qwen2}. When the visual encoder and LLM are well-aligned in terms of their representations, the LVLM can demonstrate robust multimodal capabilities, leveraging the LLM to achieve a deeper understanding of image semantics.

\subsection{Hallucination in LVLMs}
\label{sec:hallucination_in_lvlms}

Despite their effectiveness in vision-language tasks, LVLMs still face significant challenges with hallucinations. These hallucinations can be categorized into three distinct types: object, attribute, and relational hallucinations. Object hallucinations occur when models incorrectly identify objects within images. Attribute hallucinations includes state, number and action. Relational hallucinations, stem from errors in understanding spatial or contextual relationships between objects (\eg, ``above,'' ``below,'' ``left,'' or ``right''). The evaluation of these hallucinations can be broadly divided into two categories: discriminative and generative tasks. Discriminative hallucination evaluations \cite{li2023evaluating, lovenia2023negative, hu2023ciem} typically involve yes/no questions (\eg, ``Is the lemon green in this image?'') to assess hallucinations in LVLMs. Generative hallucination assessments \cite{gunjal2024detecting, liu2023mitigating, jing2023faithscore, wang2023evaluation, sun2023aligning, wang2023llm}, in contrast, evaluate hallucinations based on the generative responses produced by LVLMs.

\subsection{Detecting Hallucination in LLMs/LVLMs}

Extensive research has been dedicated to hallucination detection in LLMs \cite{arteaga2024hallucination, farquhar2024detecting, chen2024internvl, xu2023understanding, agrawal2024language}. While numerous approaches have been proposed to mitigate hallucinations in LVLMs \cite{hu2023ciem, liu2023mitigating, you2023ferret, gunjal2024detecting, zhai2023halle, lu2023evaluation, jain2023vcoder, huang2023opera, stiennon2020learning, jiang2023hallucination, bai2023qwen, bai2025qwen2, li2023monkey, chen2024internvl, leng2023mitigating, liu2024paying, wang2024mitigating}, there remains a significant gap in the literature regarding hallucination detection specifically in LVLMs. Our work bridges this gap by directly addressing hallucination detection in LVLMs. We propose that LVLMs exhibit distinct cross-modal attention patterns during hallucinations, which differ from those observed in non-hallucination cases. By analyzing the model's attention to visual tokens during the decoding of initial outputs, we develop an effective method to distinguish between hallucinated and non-hallucinated samples. 

\section{Method}

\subsection{Cross-modal Attention in LVLM}
\label{sec:definition_of_A}

We first define the cross-modal attention $\mathbf{A}$ in LVLMs. Using InstructBLIP or LLaVA as examples, which employ Q-former or MLP as their projector, we identify three key components: the CLIP visual encoder $f_\text{V}$, the projector $f_\text{P}$, and the LLM $f_\text{LLM}$. Given an image $x_i$ and a text $x_t$, the visual encoder generates visual features $f_\text{V}(x_i)$. These features are then aligned with the text by the projector, resulting in $f_\text{P}(f_\text{V}(x_i), x_t)$, which is formatted to fit the input requirements of the LLM $f_\text{LLM}$. Following common practice in some LVLMs, we also pass the text $x_t$ through the projector $f_\text{P}$. The resulting visual tokens of length $N$ are then fed into the LLM. Subsequently, the LLM produces a sequence of responses with length $M$. Let $\mathbf{A}_{m,n}(x_i, x_t)$ denote the attention assigned by the generated $m$-th token to the $n$-th image token. Different from \cite{zhang2024pip, zhang2025fighting}, we define our \textbf{cross-modal attention} $\mathbf{A}(x_i, x_t)$ as \emph{the average of the LLM's attention weights from the generated tokens to each visual token across all layers and attention heads}, \ie,
\begin{equation}
    \centering
    \label{eq:cma}
    \mathbf{A}=\mathbf{A}(x_i, x_t)=\frac{\Sigma_{m}\mathbf{A}_{m,n}(x_i, x_t)}{M}
\end{equation}
It has a shape defined by three dimensions: $(N, L, H)$, where $L$ represents the number of LLM layers, and $H$ denotes the number of multi-head attention heads. The specific dimensions of $\mathbf{A}$ vary across models. For example, LLaVA-v1.5-7B has dimensions $(N=576, L=32, H=32)$, while InstructBLIP-7B and InternVL-2.5-4B both have $(N=32, L=32, H=32)$ and $(N=256, L=36, H=16)$ respectively. Notably, the Qwen2.5-VL models feature dynamically determined image token counts $N$, which vary based on input resolution. For fixed resolutions, $N$ can be uniquely determined, for instance, an image resolution of $336\times336$ corresponds to $(N=144, L=28, H=28)$ in Qwen2.5-VL-7B.

\subsection{Cross-modal Attention Excels in Revealing Trace of LVLM Hallucinations}
\label{sec:attention_between_hallucination_and_non_hallucination}

\begin{figure}[!htbp]
  \centering
    \begin{subfigure}{\linewidth}
        \includegraphics[width=\linewidth]{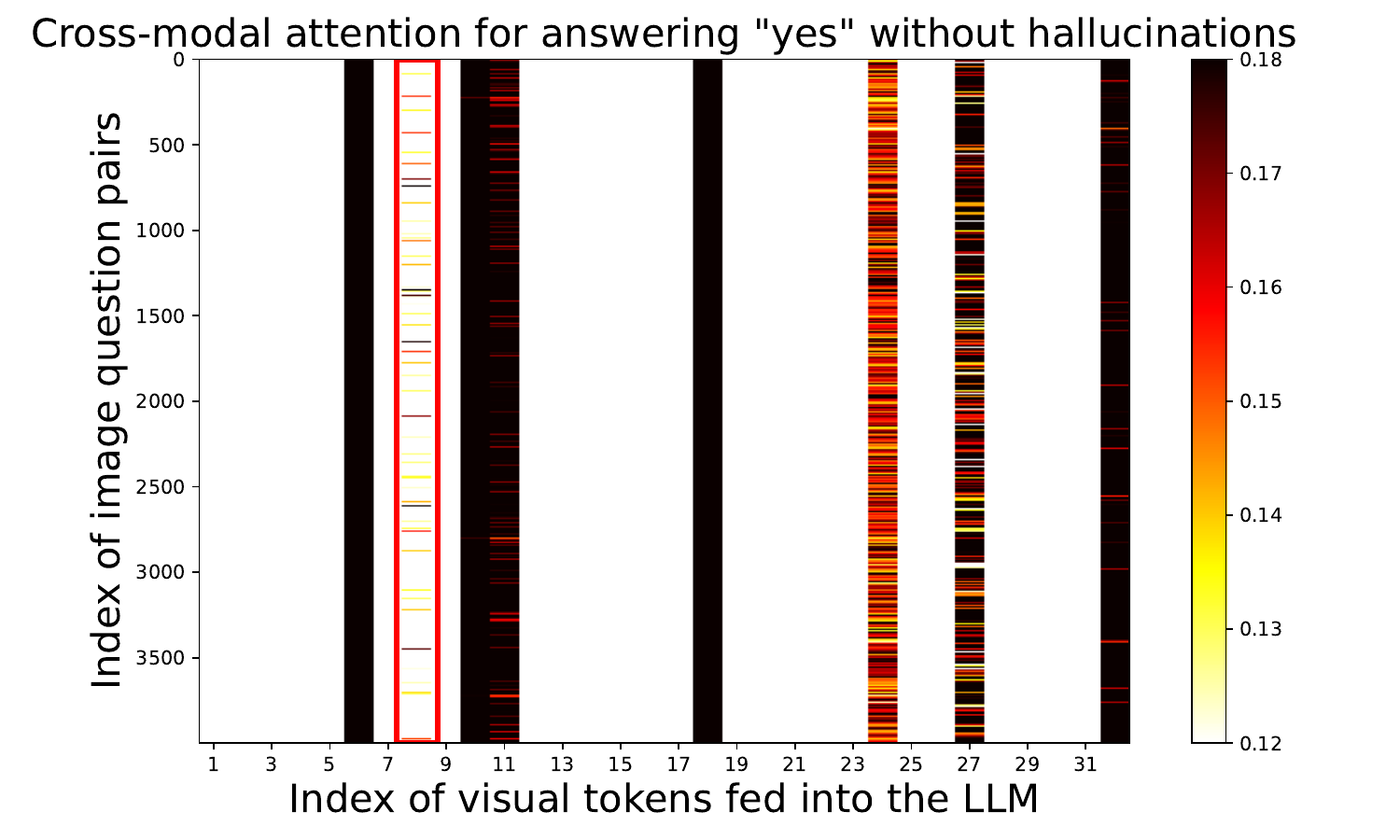}
        \caption{Cross-modal attention of $\textbf{A}^\text{T}$, \ie, answers without hallucinations.}
        \label{fig:res_true_answer_yes}
    \end{subfigure}
    \hfill
    \begin{subfigure}{\linewidth}
        \includegraphics[width=\linewidth]{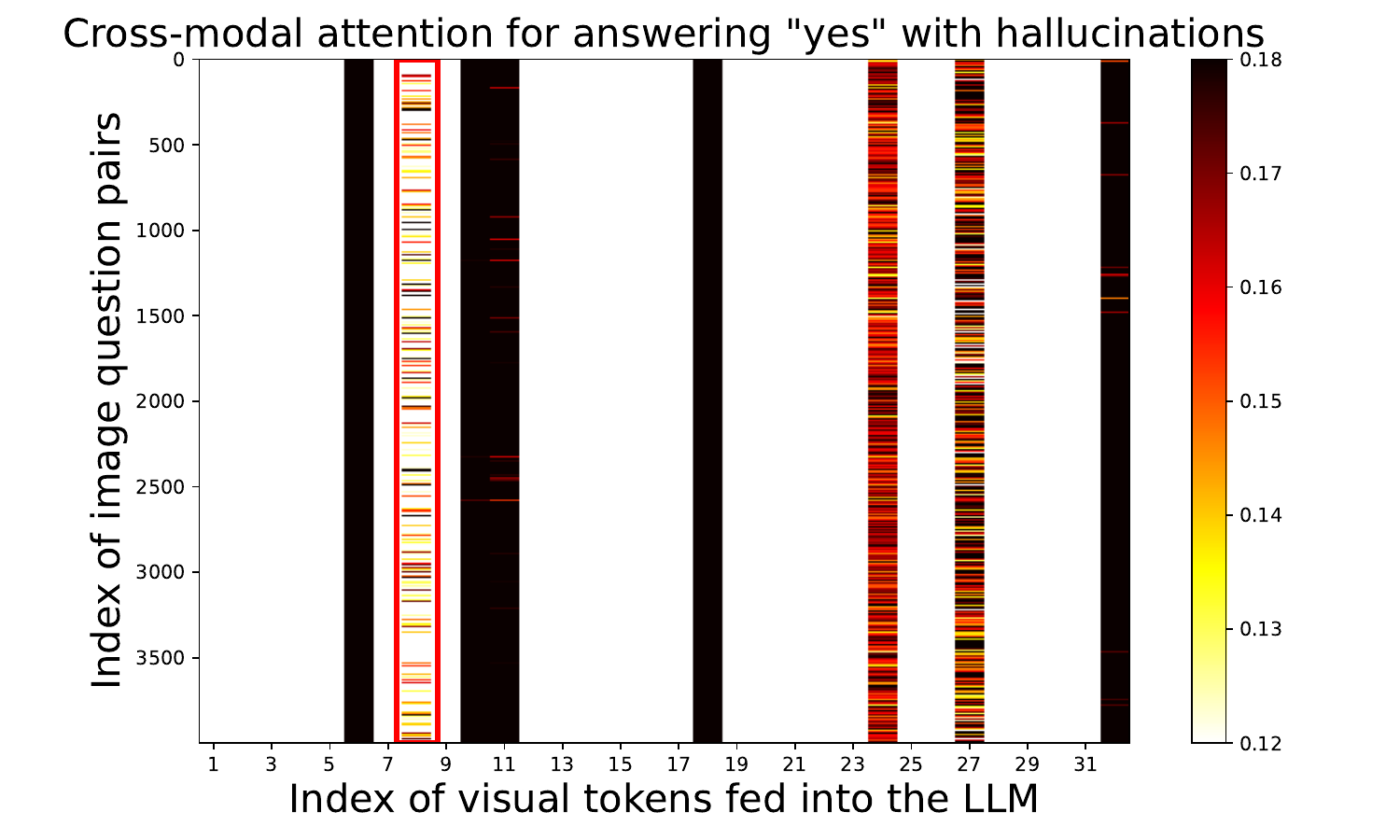}
        \caption{Cross-modal attention of $\textbf{A}^\text{H}$, \ie, answers with hallucinations.}
        \label{fig:res_false_answer_yes}
    \end{subfigure}
  \caption{The visualization of cross-modal attention.}
  \label{fig:attention_of_four_class}
\end{figure}

To investigate whether $\mathbf{A}$, as defined in \cref{eq:cma}, differs between hallucination and non-hallucination samples, we fed images from the POPE dataset \cite{li2023evaluating} into InstructBLIP-7B. These images were then categorized into two groups, $\textbf{A}^\text{T}$ (non-hallucination, \ie, truth) and $\textbf{A}^\text{H}$ (hallucination), based on whether hallucination occurred. The cross-modal attention for these groups are presented in \cref{fig:attention_of_four_class}. For visualization, we randomly selected 4,000 images from each category. Each row in the figure corresponds to an image-question pair, while each column represents a visual token. We analyzed one randomly selected layer of the LLM and extracted the maximum attention values across the multi-head attention dimension.

From \cref{fig:attention_of_four_class}, one key observations emerges: \emph{\textbf{The red highlighting reveals significant differences in cross-modal attention patterns between non-hallucination and hallucination conditions for identical responses.}} For instance, the cross-modal attention value of the 8th token was notably higher in the hallucination case (\cref{fig:res_false_answer_yes}) compared to the non-hallucination case (\cref{fig:res_true_answer_yes}). The similar differences were observed across layers and attention heads of a variety of LVLMs. 

\subsection{DHCP: Hallucination Detection Method}
\label{sec:dhcp_d}

Based on the differences between attention patterns observed in hallucinatory and non-hallucination examples (as shown in \cref{fig:attention_of_four_class}), we developed a detector designed to \emph{identify whether the model is experiencing hallucinations}. For our DHCP hallucination detector, we employ a straightforward MLP with an input of cross-modal attention $\mathbf{A}$ of shape $(N, L, H)$. The attention $\mathbf{A}$ is flattened and passed through the detector. In general, we use a two-layer MLP with a hidden layer dimension of 128 and an output dimension of 2 to represent hallucinatory and non-hallucinatory samples, respectively. Specifically, for models like InstructBLIP, which have fewer image tokens (only 32), we implement a cascade of two MLPs as the DHCP detector. In the cascaded architecture, the second MLP only processes samples identified as hallucinatory by the first MLP; if both detectors agree, the sample is confirmed as hallucinatory. For other models such as Qwen2.5-VL, InternVL-2.5, and LLaVA-v1.5, a single MLP suffices for detection. 

To train the DHCP detector effectively, we collected cross-modal attention data from hallucinatory and non-hallucinatory samples in the training set. These samples were labeled as class 0 (non-hallucinatory) and class 1 (hallucinatory), respectively. Notably, hallucinatory samples constitute only approximately 10-20\% of the total dataset, leading to a significant class imbalance. To mitigate this issue, we implemented a weighted sampling strategy during training, where sample weights were assigned in inverse proportion to their class sizes. This approach helps prevent the detector from biasing towards the majority (non-hallucinatory) class and improve the recall of hallucinatory samples.

\section{Experiment}

\subsection{Datasets}

We utilized four distinct datasets in our experiments, as follows:

\noindent
\textbf{POPE-official}. POPE \cite{li2023evaluating} is a classical LVLM object hallucination dataset that contains three clusters: Random, Popular, and Adversarial. In all three clusters, the positive example questions select three objects already present in the image. The negative example questions differ in their choices: (a) ``random'' selects three random objects not present in the image, (b) ``popular'' selects three objects not present in the image but with the highest frequency in the dataset, and (c) ``adversarial'' selects three objects not present in the image but with the most co-occurrences with objects that do exist in the image. The official POPE contains 500 images per cluster, with 3 positive and 3 negative examples per image, \ie, a total of 9,000 questions. We denote the official POPE dataset as POPE-official. Using the same methodology as POPE, we also generated POPE-COCO and POPE-GQA, which provides an effective framework for training hallucination detectors and offers a robust approach to validate our method.

\noindent
\textbf{POPE-COCO}. We identified 22,670 images from the COCO-val2014 dataset that contain at least three objects based on instance-level annotations. These images were split into training and test sets at a ratio of 8:2, ensuring all images from POPE-official were included in the test set to avoid overlap. This resulted in a training set, POPE-COCO-train, with 18,136 images, and a test set, POPE-COCO-test, with 4,534 images. Using a POPE-like approach, we generated 326,448 discriminative questions for the training set and 81,612 discriminative questions for the test set to evaluate hallucinations.

\noindent
\textbf{POPE-GQA}. We applied SEEM \cite{zou2023segment} to the GQA dataset to generate instance-level annotations. Using the same methodology as POPE, we then created discriminative questions. The POPE-GQA dataset consists of 15,000 images and 90,000 questions.

\noindent
\textbf{AMBER}. Building on POPE's evaluation of object hallucinations, AMBER \cite{wang2023llm} introduces assessments for attribute and relation hallucinations. It features 14,216 discriminative queries. We divided AMBER into AMBER-training and AMBER-test at an 8:2 ratio. To ensure no overlap, we carefully split the dataset so that each image appears only in one of the subsets. This resulted in AMBER-training with 803 images and 11,422 questions, and AMBER-test with 201 images and 2,794 questions.

\noindent
\textbf{COCO-Caption}. We extracted captions for images from COCO-train2014 and COCO-val2014 \cite{lin2014microsoft} datasets, utilizing CHAIR scores \cite{rohrbach2018object} to evaluate hallucinations. The extracted attention data were then used to form COCO-Caption-train and COCO-Caption-test.

\subsection{Experimental Settings}

\noindent
\textbf{Models}. We primarily employed four representative models to validate our DHCP: Qwen2.5-VL-7B \cite{bai2025qwen2}, InternVL-2.5-4B \cite{chen2024internvl}, LLaVA-v1.5-7B \cite{liu2024visual}, and InstructBLIP-7B \cite{dai2024instructblip}. Although these models exhibit different hallucination performance, the total number of data samples remains consistent across experiments.

\noindent
\textbf{Implementation details}. Our DHCP detector consists of a two-layer multilayer perceptron (MLP). The first layer projects cross-modal attention features into a 128-dimensional hidden space, and the second layer maps these hidden states to two output classes: hallucination and non-hallucination. We choose MLP for its effectiveness and simplicity. It is also easy to adopt more sophisticated models as our detector. We used the Adam optimizer with an initial learning rate of 0.001 and a batch size of 1024. The number of training epochs varied depending on the size of the training set. While careful hyperparameter tuning could further enhance DHCP, our focus was not on extensive optimization. 

\noindent
\textbf{DHCP training and inference costs}. DHCP demonstrates exceptionally low training costs, primarily due to its implementation of a lightweight two-layer MLP for detection. For instance, in Qwen2.5-VL, training on POPE-COCO-train containing over 300,000 samples can be completed in 30 seconds per epoch. During inference, DHCP supervises cross-modal attention within LVLMs and detecting hallucinations efficiently. The introduction DHCP two-layer MLP adds little computational overhead compared to LVLMs, which often scales to over a billion parameters.

\noindent
\textbf{Comparison with randomized baseline}. Assuming the probability of encountering a hallucinatory sample is $q$ in a dataset of 100 samples, Given that hallucinations typically occur infrequently, we established a random guessing baseline: predicting the hallucinatory class with probability $p=q$, as showon in \cref{tab:dhcp_random_guessing}.

\begin{table}[!htbp]
    \caption{Results of random guessing.}
    \resizebox{\linewidth}{!}{
    \begin{tabular}{c|c|cccc}
        \toprule
        Method & Metric & Precision & Recall & F1-score & Support \\
        \midrule
        \multirow{5}{*}{$p=q$} & Non-hallucination & $1-q$ & $1-q$ & $1-q$ & $100\times(1-q)$ \\
        & Hallucination & $q$ & $q$ & $q$ & $100\times q$ \\
        \cmidrule(lr){2-6}
        & Macro avg & $50.00$ & $50.00$ & $50.00$ & $100$ \\
        & Weighted avg & $1-2q+2q^2$ & $1-2q+2q^2$ & $1-2q+2q^2$ & $100$ \\
        \cmidrule(lr){2-6}
        & Accuracy & \multicolumn{4}{c}{$1-2q+2q^2$} \\
        \bottomrule
        \end{tabular}
    }
    \label{tab:dhcp_random_guessing}
\end{table}

\subsection{DHCP on Discriminative Tasks about Object Hallucinations with Different LVLMs}
\label{sec:dhcp_id}

We demonstrate the effectiveness of DHCP in detecting object hallucinations using POPE-COCO dataset. We extract cross-modal attention features for four LVLMs on POPE-COCO-train, including QwenVL, InternVL, and others. Using these cross-modal attentions, we then train DHCP detectors where the presence or absence of hallucinations serves as the label. Subsequently, we apply the trained DHCP detector to analyze cross-modal attention from POPE-COCO-test for hallucination detection, with results presented in \cref{tab:dhcp_pope_detect_coco}. The experimental results show that our method achieves impressive recall rates of approximately 85\% for hallucination samples while maintaining precision levels between 60\% to 80\% across all four LVLMs. This clearly demonstrates the robust performance of the DHCP detector in identifying hallucinations. 

\begin{table}[!htbp]
    \caption{Results of DHCP hallucination detection on POPE-COCO-test. The detector is trained on POPE-COCO-train.}
    \resizebox{\linewidth}{!}{
    \begin{tabular}{c|c|cccc}
        \toprule
        LVLMs & Metric & Precision & Recall & F1-score & Support \\
        \midrule
        \multirow{5}{*}{Qwen2.5-VL-7B} & Non-hallucination & 98.25 & 98.07 & 98.16 & 71414 \\
        & Hallucination & 86.65 & 87.73 & 87.19 & 10198 \\
        \cmidrule(lr){2-6}
        & Macro avg & 92.45 & 92.90 & 92.67 & 81612 \\
        & Weighted avg & 96.80 & 96.78 & 96.79 & 81612 \\
        \cmidrule(lr){2-6}
        & Accuracy & \multicolumn{4}{c}{96.78} \\
        \midrule
        \multirow{5}{*}{LLaVA-v1.5-7B} & Non-hallucination & 97.88 & 97.27 & 97.57 & 71003 \\
        & Hallucination & 82.46 & 85.87 & 84.13 & 10609 \\
        \cmidrule(lr){2-6}
        & Macro avg & 90.17 & 91.57 & 90.85 & 81612 \\
        & Weighted avg & 95.87 & 95.79 & 95.82 & 81612 \\
        \cmidrule(lr){2-6}
        & Accuracy & \multicolumn{4}{c}{95.79} \\
        \midrule
        \multirow{5}{*}{InternVL-2.5-4B} & Non-hallucination & 98.54 & 94.34 & 96.40 & 72423 \\
        & Hallucination & 66.62 & 88.98 & 76.19 & 9189 \\
        \cmidrule(lr){2-6}
        & Macro avg & 82.58 & 91.66 & 86.30 & 81612 \\
        & Weighted avg & 94.95 & 93.74 & 94.12 & 81612 \\
        \cmidrule(lr){2-6}
        & Accuracy & \multicolumn{4}{c}{93.74} \\
        \midrule
        \multirow{5}{*}{InstructBLIP-7B} & Non-hallucination & 96.30 & 92.77 & 94.50 & 70062 \\
        & Hallucination & 64.13 & 78.41 & 70.55 & 11550 \\
        \cmidrule(lr){2-6}
        & Macro avg & 80.22 & 85.59 &  82.53 & 81612 \\
        & Weighted avg & 91.75 & 90.74 & 91.12 & 81612 \\
        \cmidrule(lr){2-6}
        & Accuracy & \multicolumn{4}{c}{90.74} \\
        \bottomrule
        \end{tabular}
    }
    \label{tab:dhcp_pope_detect_coco}
\end{table}

\subsection{DHCP on Discriminative Tasks about Other Hallucination Types and Datasets}
\label{sec:dhcp_amber}

\begin{table}[!htbp]
    \caption{Results of DHCP hallucination detection on AMBER-test. The detector is trained on AMBER-train.}
    \resizebox{\linewidth}{!}{
    \begin{tabular}{c|c|cccc}
        \toprule
        LVLMs & Metric & Precision & Recall & F1-score & Support \\
        \midrule
        \multirow{5}{*}{Qwen2.5-VL-7B} & Non-hallucination & 93.24 & 93.74 & 93.49 & 2444 \\
        & Hallucination & 54.60 & 52.57 & 53.57 & 350 \\
        \cmidrule(lr){2-6}
        & Macro avg & 73.92 & 73.16 & 73.53 & 2794\\
        & Weighted avg & 88.40 & 88.58 & 88.49 & 2794\\
        \cmidrule(lr){2-6}
        & Accuracy & \multicolumn{4}{c}{88.58} \\
        \midrule
        \multirow{5}{*}{LLaVA-v1.5-7B} & Non-hallucination & 89.39 & 86.14 & 87.74 & 2280 \\
        & Hallucination & 46.98 & 54.58 & 50.50 & 513 \\
        \cmidrule(lr){2-6}
        & Macro avg & 68.19 & 70.36 & 69.12 & 2793 \\
        & Weighted avg & 81.60 & 80.34 & 80.90 & 2793 \\
        \cmidrule(lr){2-6}
        & Accuracy & \multicolumn{4}{c}{80.34} \\
        \midrule
        \multirow{5}{*}{InternVL-2.5-4B} & Non-hallucination & 90.38 & 80.97 & 85.42 & 2228 \\
        & Hallucination & 46.73 & 65.96 & 54.71 & 564 \\
        \cmidrule(lr){2-6}
        & Macro avg & 68.56 & 73.46 & 70.06 & 2792 \\
        & Weighted avg & 81.56 & 77.94 & 79.21 & 2792 \\
        \cmidrule(lr){2-6}
        & Accuracy & \multicolumn{4}{c}{77.94} \\
        \midrule
        \multirow{5}{*}{InstructBLIP-7B} & Non-hallucination & 93.06 & 80.66 & 86.42 & 2228 \\
        & Hallucination & 49.94 & 76.24 & 60.35 & 564\\
        \cmidrule(lr){2-6}
        & Macro avg & 71.50 & 78.45 & 73.38 & 2792 \\
        & Weighted avg & 84.35 & 79.76 & 81.15 & 2792 \\
        \cmidrule(lr){2-6}
        & Accuracy & \multicolumn{4}{c}{79.76} \\
        \bottomrule
        \end{tabular}
        
    }
    \label{tab:dhcp_pope_detect_amber}
\end{table}

\begin{table}[!htbp]
    \caption{Results of DHCP detectors on AMBER-test per-subtype hallucinations on InstructBLIP-7B.}
    \label{tab:dhcp_pope_detect_amber_subtype}
    \resizebox{\linewidth}{!}{
    \begin{tabular}{c|c|cccc}
        \toprule
        AMBER-test & \multirow{2}{*}{Metric} & \multirow{2}{*}{Precision} & \multirow{2}{*}{Recall} & \multirow{2}{*}{F1-score} & \multirow{2}{*}{Support} \\
        subtype & \\
        \midrule
        \multirow{5}{*}{\shortstack{Attribute}} & Non-hallucination & 88.40 & 77.43 & 82.55 & 1161 \\
        & Hallucination & 42.42 & 62.06 & 50.39 & 311 \\
        \cmidrule(lr){2-6}
        & Macro avg & 65.41 & 69.75 & 66.47 & 1472 \\
        & Weighted avg & 78.68 & 74.18 & 75.76 & 1472 \\
        \cmidrule(lr){2-6}
        & Accuracy & \multicolumn{4}{c}{74.18} \\
        \midrule
        \multirow{5}{*}{Object} & Non-hallucination & 99.18 & 99.30 & 99.24 & 854 \\
        & Hallucination & 95.45 & 9474 & 95.09 & 133 \\
        \cmidrule(lr){2-6}
        & Macro avg & 97.32 & 97.02 & 97.17 & 987 \\
        & Weighted avg & 98.68 & 98.68 & 98.68 & 987 \\
        \cmidrule(lr){2-6}
        & Accuracy & \multicolumn{4}{c}{98.68} \\
        \midrule
        \multirow{5}{*}{Relation} & Non-hallucination & 84.75 & 23.47 & 36.76 & 213 \\
        & Hallucination & 40.51 & 92.50 & 56.35 & 120 \\
        \cmidrule(lr){2-6}
        & Macro avg & 62.63 & 57.99 & 46.55 & 333 \\
        & Weighted avg & 68.81 & 48.35 & 43.82 & 333 \\
        \cmidrule(lr){2-6}
        & Accuracy & \multicolumn{4}{c}{48.35} \\
        \bottomrule
        \end{tabular}
    }
\end{table}

While POPE is effective for evaluating object hallucinations, it does not assess attribute and relation hallucinations. To comprehensively evaluate DHCP's ability to detect diverse types of hallucinations, including both attribute and relation hallucinations, we employ AMBER, which evaluates three categories of hallucinations (object, attribute, and relationship) through discriminative questions. 

Training DHCP on the AMBER dataset presents two key challenges: (1) The diversity of hallucination types in AMBER extends beyond simple object hallucinations, incorporating more complex scenarios \eg attribute and relation. (2) Compared to larger datasets like POPE-COCO-train, which contain tens of thousands of images and hundreds of thousands of questions, the AMBER-train dataset is relatively small, with only 1,000 images and 14,000 questions. This smaller scale may pose challenges for training DHCP effectively.
     
Despite these challenges, we successfully trained the DHCP detector on AMBER-train and evaluated its performance on the AMBER-test dataset. The experimental results across multiple LVLMs are presented in \cref{tab:dhcp_pope_detect_amber}. Even when faced with more complex hallucination types and a smaller training dataset, DHCP demonstrates robust hallucination detection performance. It successfully recalls over 50\% of hallucinations with 50\% precision across multiple LVLMs, showcasing its strong potential for practical applications.

Interestingly, AMBER exhibits an uneven distribution across different types of hallucinations. While \Cref{tab:dhcp_pope_detect_amber} presents the overall performance of hallucination detection on the AMBER-test set, it does not provide insights into how DHCP performs on individual hallucination types. This raises a critical question regarding whether DHCP's effectiveness might be limited to specific types of hallucinations. To address this concern, we analyze DHCP's detection performance across different hallucination types in the AMBER-test set. \Cref{tab:dhcp_pope_detect_amber_subtype} demonstrates that DHCP exhibits robust performance on the subtype of AMBER-test, achieving F1 scores of approximately 90 for object hallucinations and around 50 for attribute and relation hallucinations. Notably, despite encountering a greater variety of hallucination types and more sparse training data, DHCP still demonstrates strong hallucination detection capabilities.

\subsection{DHCP on Discriminative Tasks about Out-of-distribution Datasets}
\label{sec:dhcp_ood}

In \cref{sec:dhcp_id}, we evaluate the DHCP detector's performance on the POPE-COCO dataset. Although the DHCP detector was trained on POPE-COCO-train and tested on POPE-COCO-test, the fact that all images originate from the COCO dataset raises concerns about whether the DHCP detector is merely fitting to the COCO image distribution rather than developing generalized hallucination detection capabilities. To address these concerns and validate the generalization ability of the DHCP detector for out-of-distribution data, we tested its performance on the POPE-GQA-test dataset. As shown in \cref{tab:dhcp_pope_detect_gqa}, when the image distributions differ (from COCO to GQA), the hallucination detection performance of DHCP degrades but remains robust, achieving over 60\% recall on hallucinated samples and approximately 0.7 F1 score across all samples. These initial results demonstrate the potential for improving DHCP's detection performance by incorporating additional hallucination samples from diverse datasets or leveraging synthetic data. By expanding the training set with more hallucination examples from various domains or from the extensive synthetic data, we may further enhance the generalization and detection capabilities of DHCP.

\begin{table}[!htbp]
    \caption{Results of DHCP hallucination detection on POPE-GQA-test. The detector is trained on POPE-COCO-train.}
    \resizebox{\linewidth}{!}{
    \begin{tabular}{c|c|cccc}
        \toprule
        LVLMs & Metric & Precision & Recall & F1-score & Support \\
        \midrule
        \multirow{5}{*}{Qwen2.5-VL-7B} & Non-hallucination & 95.57 & 89.39 & 92.38 & 80499 \\
        & Hallucination & 41.93 & 64.92 & 50.95 & 9501 \\
        \cmidrule(lr){2-6}
        & Macro avg & 68.75 & 77.15 & 71.66 & 90000 \\
        & Weighted avg & 89.91 & 86.81 & 88.00 & 90000 \\
        \cmidrule(lr){2-6}
        & Accuracy & \multicolumn{4}{c}{86.81} \\
        \midrule
        \multirow{5}{*}{LLaVA-v1.5-7B} & Non-hallucination & 94.41 & 88.33 & 91.26 & 80415 \\
        & Hallucination & 36.41 & 56.09 & 44.16 & 9585 \\
        \cmidrule(lr){2-6}
        & Macro avg & 65.41 & 72.21 & 67.71 & 90000 \\
        & Weighted avg & 88.23 & 84.89 & 86.25 & 90000 \\
        \cmidrule(lr){2-6}
        & Accuracy & \multicolumn{4}{c}{84.89} \\
        \midrule
        \multirow{5}{*}{InternVL-2.5-4B} & Non-hallucination & 96.75 & 85.00 & 90.50 & 81318 \\
        & Hallucination & 34.28 & 73.27 & 46.71 & 8682 \\
        \cmidrule(lr){2-6}
        & Macro avg & 65.51 & 79.13 & 68.60 & 90000 \\
        & Weighted avg & 90.72 & 83.87 & 86.27 & 90000 \\
        \cmidrule(lr){2-6}
        & Accuracy & \multicolumn{4}{c}{83.87} \\
        \midrule
        \multirow{5}{*}{InstructBLIP-7B} & Non-hallucination & 96.89 & 75.73 & 85.01 & 78553 \\
        & Hallucination & 33.35 & 83.32 & 47.63 & 11447 \\
        \cmidrule(lr){2-6}
        & Macro avg & 65.12 & 79.53 & 66.32 & 90000 \\
        & Weighted avg & 88.81 & 76.70 & 80.26 & 90000 \\
        \cmidrule(lr){2-6}
        & Accuracy & \multicolumn{4}{c}{76.70} \\
        \bottomrule
        \end{tabular}
        
    }
    \label{tab:dhcp_pope_detect_gqa}
\end{table}

\subsection{DHCP on Generative Image Captioning Task}
\label{sec:coco_caption}

In previous \cref{sec:dhcp_id,sec:dhcp_ood,sec:dhcp_amber}, we examined in detail discriminative hallucinations such as POPE and AMBER. We now extend our investigation to the realm of generative hallucinations by applying DHCP detection methods. To achieve this, we generated image captions using the COCO-Caption-train dataset and assessed the generated captions for hallucinations through CHAIR \cite{rohrbach2018object}. We utilize these cross-modal attention patterns and corresponding labels to develop DHCP detectors capable of identifying generative hallucinations. As \cref{tab:mlp_caption_test} highlights, our approach demonstrates promising results across various models, achieving F1 scores of 75 for LLaVA and InstructBLIP, and 45 for QwenVL and InternVL.

\begin{table}[!htbp]
    \caption{Results of DHCP hallucination detection on COCO-Caption-test. The detector is trained on COCO-Caption-train.}
    \label{tab:mlp_caption_test}
    \resizebox{\linewidth}{!}{
    \begin{tabular}{c|c|cccc}
        \toprule
        LVLMs & Metric & Precision & Recall & F1-score & Support \\
        \midrule
        \multirow{5}{*}{Qwen2.5-VL-7B} & Non-hallucination & 82.97 & 77.85 & 80.33 & 30765 \\
        & Hallucination & 41.44 & 49.51 & 45.12 & 9739 \\
        \cmidrule(lr){2-6}
        & Macro avg & 62.21 & 63.68 & 62.73 & 40504 \\
        & Weighted avg & 72.98 & 71.04 & 71.86 & 40504 \\
        \cmidrule(lr){2-6}
        & Accuracy & \multicolumn{4}{c}{71.04} \\
        \midrule
        \multirow{5}{*}{LLaVA-v1.5-7B} & Non-hallucination & 74.73 & 70.58 & 72.60 & 19520 \\
        & Hallucination & 73.98 & 77.80 & 75.84 & 20984 \\
        \cmidrule(lr){2-6}
        & Macro avg & 74.36 & 74.19 & 74.22 & 40504 \\
        & Weighted avg & 74.34 & 74.32 & 74.28 & 40504 \\
        \cmidrule(lr){2-6}
        & Accuracy & \multicolumn{4}{c}{74.32} \\
        \midrule
        \multirow{5}{*}{InternVL-2.5-4B} & Non-hallucination & 78.78 & 73.99 & 76.31 & 29088 \\
        & Hallucination & 42.61 & 49.21 & 45.67 & 11416 \\
        \cmidrule(lr){2-6}
        & Macro avg & 60.69 & 61.60 & 60.99 & 40504 \\
        & Weighted avg & 68.58 & 67.01 & 67.67 & 40504\\
        \cmidrule(lr){2-6}
        & Accuracy & \multicolumn{4}{c}{67.01} \\
        \midrule
        \multirow{5}{*}{InstructBLIP-7B} & Non-hallucination & 79.31 & 75.80 & 77.51 & 21488 \\
        & Hallucination & 73.95 & 77.66 & 75.76 & 19016 \\
        \cmidrule(lr){2-6}
        & Macro avg & 76.63 & 76.73 & 76.64 & 40504 \\
        & Weighted avg & 76.80 & 76.67  & 76.69 & 40504 \\
        \cmidrule(lr){2-6}
        & Accuracy & \multicolumn{4}{c}{76.67} \\
        \bottomrule
        \end{tabular}
    }
\end{table}

\begin{figure}[!htbp]
  \centering
  \includegraphics[width=0.5\linewidth]{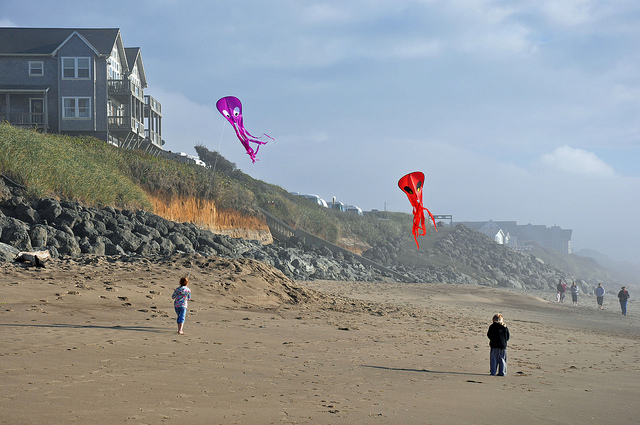}
  \caption{The example image to demonstrate the feasibility of token-by-token DHCP detection of hallucinations.}
  \label{fig:demo_caption}
\end{figure}

In our DHCP approach, the cross-modal attention is computed by averaging across the sequence of generated tokens. This method could dilute the impact of hallucinatory tokens, as their effects are dispersed throughout the entire sequence. For instance, in a 50-token caption where only 2 tokens are hallucinations, these hallucination tokens would be averaged together with the remaining 48 valid tokens during cross-modal attention calculation. A more sophisticated approach would involve evaluating cross-modal attention on a per-token basis rather than averaging across the token sequence dimension. To investigate this, we conducted a simple test: while still employing cross-modal attention training averaged over the sequence dimension, we examined the cross-modal attention for each individual token. As illustrated in \cref{fig:demo_caption}, the generated caption reads: ``... two kites are prominently visible: one shaped like an octopus and the other another resembling \textcolor{red!70}{a bird} or dragon.'' We highlighted in red the tokens identified as hallucinations by DHCP. Notably, even though cross-modal attention was averaged during training, our DHCP detector demonstrated the capability to identify hallucinations at the token level. This suggests that future optimizations for generation tasks could be achieved by implementing token-by-token training, where each generated token is evaluated based on its specific cross-modal attention and its classification as either hallucinatory or non-hallucinatory. This refinement will be explored in our future work.

\subsection{False Positives of DHCP are Also Risky}

\Cref{tab:dhcp_pope_detect_coco} shows that on the hallucination category, DHCP achieves an precision of 65-85\%, with the remaining samples being false alarms, \ie, non-hallucinating samples misclassified as hallucinations by DHCP. Our further analysis reveals that \emph{\textbf{a significant portion of these false alarm samples are actually ``unconfident'' samples, which warrant careful consideration}}. For the discriminative Yes/No questions, the LVLM model has a 50\% chance of randomly guessing the correct ``yes'' or ``no'' even when hallucinating. Therefore, correct answers do not necessarily indicate that the LVLM truly understood the question and responded without hallucination. Our investigation shows that samples for which our detector ``falsely'' alarms hallucinations are more likely to be cases where the model happened to guess correctly, which remain risky.

\begin{figure}[!htbp]
  \centering
    \includegraphics[width=\linewidth]{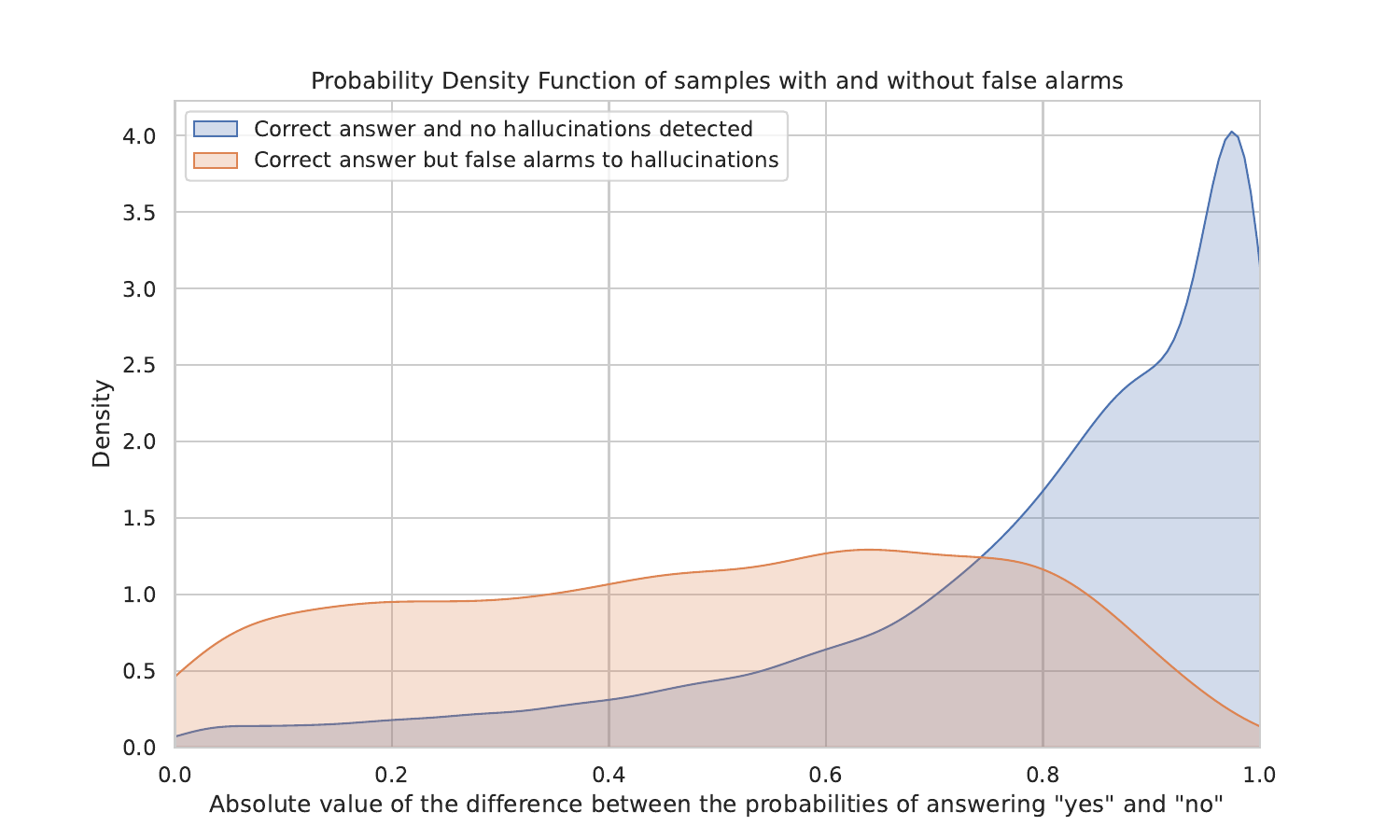}
    \caption{The probability density function of the absolute differences in probabilities between ``yes'' and ``no'' responses in LVLM. Samples flagged as false alarms by DHCP exhibit a higher tendency towards random guessing, which should also be carefully re-examined in practical applications.}
    \label{fig:pdf_of_false_alarm}
\end{figure}

To verify this, we use the absolute difference between the probabilities of LVLMs answering ``yes'' and ``no'' as a measure of the model's confidence. A larger value indicates higher confidence, while a smaller value suggests lower confidence and a tendency to guess. We classify the 27,204 samples from POPE-COCO-test-popular via DHCP and obtain the false alarm hallucination samples. We measure the average absolute difference between ``yes'' and ``no'' responses for (a) the 2,169 correctly answered but falsely alerted samples, and (b) the 21,128 correctly answered samples judged as non-hallucinating (control group). The results are 0.487 and 0.792, respectively, indicating that the ``falsely-detected'' hallucination samples are also suspicious and should be double-checked in practical applications. In \cref{fig:pdf_of_false_alarm}, we plot the distribution of these values for both sample types, and the significant gap further supports our hypothesis. This confirms that DHCP is effective for hallucination detection in LVLMs, performing better than the metrics suggest.

\begin{table}[!htbp]
    \caption{Result of classifying three hallucination types in POPE-COCO-test. This table categorizes the types of hallucinations instead of the presence or absence of hallucinations.}
    \label{tab:mlp_3_class_of_no_existing_type_test}
    \resizebox{\linewidth}{!}{
    \begin{tabular}{c|cccc}
        \toprule
        Hallucination type & Precision & Recall & F1-score & Support \\
        \midrule
        Random & 66.12 & 60.35 & 63.10 & 401 \\
        Popular & 55.73 & 96.64 & 70.70 & 1222 \\
        Adversarial & 87.62 & 47.28 & 61.42 & 1872 \\
        \midrule
        Macro avg & 69.83 & 68.09 & 65.07 & 3495 \\
        Weighted avg & 74.01 & 66.04 & 64.85 & 3495 \\
        \midrule
        Accuracy & \multicolumn{4}{c}{66.04} \\
        \bottomrule
        \end{tabular}
    }
\end{table}

\begin{table*}[!htbp]
    \centering
    \caption{The hallucination mitigation results of DHCP on POPE-official. The detector is trained on POPE-COCO-train.}
    \label{tab:reverse_pope_test}
    \resizebox{0.89\linewidth}{!}{
    \begin{tabular}{c|c|c|ccc|ccc|c|c}
        \toprule
        \multirow{2}{*}{LVLM} & POPE-official & \multirow{2}{*}{Method} & \multicolumn{3}{c|}{Label of ``Yes''} & \multicolumn{3}{c|}{Label of ``No''} & POPE & POPE \\
        \cmidrule(lr){4-6}
        \cmidrule(lr){7-9}
         & sub-dataset & & Precision & Recall & F1-score & Precision & Recall & F1-score & F1-score & Accuracy \\
        \midrule
        \multirow{6}{*}{Qwen2.5-VL-7B} & \multirow{2}{*}{Random} & Baseline & 98.88 & 76.80 & 86.45 & 81.04 & 99.13 & 89.18 & 87.81 & 87.97 \\
        & & DHCP & 96.53 & 96.53 & 96.53 & 96.53 & 96.53 & 96.53 & 96.53 & 96.53 \\
        \cmidrule(lr){2-11}
        & \multirow{2}{*}{Popular} & Baseline & 95.68 & 76.80 & 85.21 & 80.62 & 96.53 & 87.86 & 86.54 & 86.67 \\
        & & DHCP & 98.97 & 96.53 & 97.74 & 96.62 & 99.00 & 97.79 & 97.77 & 97.77 \\
        \cmidrule(lr){2-11}
        & \multirow{2}{*}{Adversarial} & Baseline & 92.46 & 76.80 & 83.90 & 80.16 & 93.73 & 86.42 & 85.16 & 85.27 \\
        & & DHCP & 95.51 & 96.53 & 96.02 & 96.50 & 95.47 & 95.98 & 96.00 & 96.00 \\
        \midrule
        \multirow{6}{*}{LLaVA-v1.5-7B} & \multirow{2}{*}{Random} & Baseline & 97.30 & 79.20 & 87.32 & 82.46 & 97.80 & 89.48 & 88.40 & 88.50\\
        & & DHCP & 95.18 & 96.07 & 95.62 & 96.03 & 95.13 & 95.58 & 95.60 & 95.60 \\
        \cmidrule(lr){2-11}
        & \multirow{2}{*}{Popular} & Baseline & 94.59 & 79.20 & 86.21 & 82.11 & 95.47 & 88.29 & 87.25 & 87.33 \\
        & & DHCP & 98.83 & 96.07 & 97.43 & 96.17 & 98.87 & 97.50 & 97.47 & 97.47 \\
        \cmidrule(lr){2-11}
        & \multirow{2}{*}{Adversarial} & Baseline & 90.00 & 79.20 & 84.26 & 81.43 & 91.20 & 86.04 & 85.15 & 85.20 \\
        & & DHCP & 94.49 & 96.07 & 95.27 & 96.00 & 94.40 & 95.19 & 95.23 & 95.23 \\
        \midrule
        \multirow{6}{*}{InternVL-2.5-4B} & \multirow{2}{*}{Random} & Baseline & 96.68 & 89.33 & 92.86 & 90.09 & 96.93 & 93.38 & 93.12 & 93.13 \\
        & & DHCP & 93.28 & 92.47 & 92.87 & 92.53 & 93.33 & 92.93 & 92.90 & 92.90 \\
        \cmidrule(lr){2-11}
        & \multirow{2}{*}{Popular} & Baseline & 88.10 & 89.33 & 88.71 & 89.18 & 87.93 & 88.55 & 88.63 & 88.63 \\
        & & DHCP & 95.52 & 92.47 & 93.97 & 92.70 & 95.67 & 94.16 & 94.07 & 94.07 \\
        \cmidrule(lr){2-11}
        & \multirow{2}{*}{Adversarial} & Baseline & 83.08 & 89.33 & 86.09 & 88.46 & 81.80 & 85.00 & 85.55 & 85.57 \\
        & & DHCP & 89.83 & 92.47 & 91.13 & 92.24 & 89.53 & 90.87 & 91.00 & 91.00 \\
        \midrule
        \multirow{6}{*}{InstructBLIP-7B} & \multirow{2}{*}{Random} & Baseline & 96.35 & 79.13 & 86.90 & 82.30 & 97.00 & 89.05 & 87.98 & 88.07 \\
        & & DHCP & 94.88 & 87.67 & 91.13 & 88.54 & 95.27 & 91.78 & 91.46 & 91.47 \\
        \cmidrule(lr){2-11}
        & \multirow{2}{*}{Popular} & Baseline & 90.47 & 79.13 & 84.42 & 81.46 & 91.67 & 86.26 & 85.34 & 85.40 \\
        & & DHCP & 95.43 & 87.67 & 91.38 & 88.59 & 95.80 & 92.06 & 91.72 & 91.73 \\
        \cmidrule(lr){2-11}
        & \multirow{2}{*}{Adversarial} & Baseline & 86.52 & 79.13 & 82.66 & 80.77 & 87.67 & 84.08 & 83.37 & 83.40 \\
        & & DHCP & 85.28 & 87.67 & 86.46 & 87.31 & 84.87 & 86.07 & 86.27 & 86.27 \\
        \bottomrule
        \end{tabular}
    }
\end{table*}

\subsection{In-depth Analysis on Different Causes of LVLM Hallucinations}
\label{sec:triple_mlp}

The POPE dataset contains three distinct clusters of negative examples: random, popular, and adversarial. These clusters suggest different potential causes for the phenomenon of ``observing something out of nothing'' hallucinations: (a) The ``popular'' cluster may stem from \emph{\textbf{popularity biases inherent in model training}}. (b) The ``adversarial'' cluster could originate from \emph{\textbf{co-occurrence biases present in both multimodal data and large language models}}. (c) The origins of the ``random'' cluster remain less clear.

To investigate these hypotheses, we developed a three-class classifier to distinguish between the different types of hallucinations. This classifier was trained using 13,824 hallucination images from the POPE-COCO-train dataset and evaluated on 3,495 hallucination images from the POPE-COCO-test dataset. As shown in \cref{tab:mlp_3_class_of_no_existing_type_test}, while the detection results were informative, they were not perfect. We attribute this imperfection to potential overlaps between the three clusters—random, popular, and adversarial. For instance, approximately 58\% of adversarial samples also belong to the popular category (\ie, they represent the five most frequently occurring objects). Similarly, random negative examples often overlap with both popular and adversarial cases. These overlaps among clusters inevitably complicate type categorization and may be mitigated by reducing such overlaps during dataset construction. Our findings provide insights into the underlying mechanisms, causes, and potential solutions for hallucinations in large vision-language models.

In summary, the ability of cross-modal attention to distinguish between different types of hallucinations (each arising from distinct causes) suggests that hallucinations with different origins exhibit distinct patterns in cross-modal attention. This observation reinforces the connection between hallucinations and cross-modal attention while highlighting the need for further research into the root causes of hallucinations in future studies.

\subsection{Exploration on Hallucination Mitigation via DHCP for Discriminative Yes/No Tasks}
\label{sec:dhcp_mitigate}

DHCP primarily focuses on hallucination detection. However, in discriminative Yes/No tasks, hallucinations can be effectively mitigated by simply flipping the answers identified as hallucinations through DHCP detection. We applied this approach to POPE-official and present the results in \cref{tab:reverse_pope_test}. Notably, even with a simple yet lightweight hallucination detector, DHCP achieved improvements across multiple LVLMs evaluated on the POPE-official benchmarks. Importantly, our DHCP methodology places its primary emphasis on hallucination detection; hallucination mitigation, in this context, are merely incidental to the core discrimination task.

\section{Conclusion}
\label{sec:conclusion}

In this paper, we propose a hallucination detection method called DHCP, which leverages the cross-modal attention mechanisms of LVLMs. Despite its simplicity and without requiring additional training or inference for LVLMs, DHCP demonstrates effective hallucination detection capabilities across multiple benchmarks, including POPE, COCO, GQA, AMBER, as well as tasks such as VQA and image captioning. Extensive experiments conducted on Qwen2.5-VL, InternVL-2.5, LLaVA-v1.5, and InstructBLIP validate the robustness of our approach. When integrated with our efficient DHCP framework, LVLMs gain the ability to identify hallucinations during inference, thereby reducing their potential risks. Remarkably, DHCP provides an almost zero-cost solution for enhancing the reliability of LVLMs, offering significant improvements without additional computational overhead or training requirements.
Future work will investigate DHCP at the token level, aiming to build a finer-grained LVLM hallucination detector that could precisely discover the exact place where the hallucination happens.

\clearpage
\newpage

\begin{acks}
This work was supported by Young Elite Scientists Sponsorship Program by CAST (2023QNRC001), the National Key R\&D Program of China (2023YFB4502200), the National Natural Science Foundation of China (No. 62376024, 62325405, 62104128, 62203257, 62031017, 62406159, U21B2031), Tsinghua University Initiative Scientific Research Program,  Beijing National Research Center for Information Science, Technology (No. BNR2024TD03001), Beijing Innovation Center for Future Chips, and State Key laboratory of Space Network and Communications.
\end{acks}

\bibliographystyle{ACM-Reference-Format}
\balance
\bibliography{sample-base}

\clearpage
\newpage

\appendix

\section{Novelty of DHCP}
\label{sec:novelty}

We summarise our key innovations as follows: 

(1) Our primary objective is to detect hallucinations across a wide range of practical applications of LVLMs. Focusing on both efficiency and effectiveness, we avoided approaches based on large model inference such as contrast decoding methods (\eg, VCD, ICD, \etc) or backtracking-based approaches (\eg, OPERA), which require multiple generation steps. Instead, we approached hallucination detection from a novel perspective: can hallucination-specific features be identified through internal states and intermediate values within LVLMs? This enables almost ``zero-cost'' hallucination detection. Our investigation revealed that cross-modal attention serves as a critical indicator for distinguishing hallucinations, which is promising to be adopted in real-world LVLMs.

(2) While differences in cross-modal attention between hallucinated and non-hallucinated samples were observed, several key technical design decisions were necessary to operationalise this approach. We should highlight that it is nontrivial to detect LVLM hallucinations based on our CMA.
For instance, we explored which specific tokens of cross-modal attention should be utilized throughout the generation process (token by token). Through extensive explorations and experiments, we determined that averaging cross-modal attention across all generated tokens provides effective results for hallucination detection. Furthermore, after evaluating various detector architectures, we found a simple two-layer MLP to be sufficiently robust for this task, which is astonishing and valuable especially in the era of LLMs. Additional explorations into training details and our practical usages, such as sample balancing and hallucination mitigation, further contributed to the effectiveness of our approach. 

(3) Despite its simplicity, DHCP demonstrates strong performance in hallucination detection across diverse models (including QwenVL, InternVL, LLaVA, InstructBLIP), datasets (such as POPE-COCO and AMBER), and tasks like visual question answering (VQA) and image captioning. Notably, DHCP maintains robust performance even on the out-of-distribution POPE-GQA dataset, highlighting its strong generalisability.

\section{Addition Experiment of DHCP on Out-of-distribution Datasets}
\label{sec:addition_ood_experiment}

Although we evaluated out-of-distribution data in \cref{sec:dhcp_ood,tab:mlp_caption_test}, there are still concerns about out-of-distribution generalization of DHCP. To this end, we included the following experiments to demonstrate the generalization of DHCP on out-of-distribution data in discriminative and generative tasks.

We constructed POPE-ImageNet using the same methodology as POPE, leveraging validation images from ImageNet. Notably, we selected ImageNet as our test set rather than COCO due to its relative independence and reduced risk of dataset overlap, making it more suitable for evaluating out-of-distribution (OOD) generalization. For training, we utilized two distinct datasets: the first consisted solely of POPE-COCO, while the second encompassed a broader combination of POPE-COCO, POPE-GQA, and POPE-VG100K. We conducted experiments using Qwen2.5-VL-7B. \Cref{tab:dhcp_pope_imagenet} shows two important conclusions: (1) DHCP demonstrates robust detection capabilities on OOD POPE-ImageNet dataset when trained on COCO+GQA+VG100K, highlighting its OOD generalization performance. (2) Incorporating diverse training datasets significantly enhances DHCP. In real-world applications, we can collect sufficient and diverse training data, which can significantly improve performance and ensure the generalizability and practicality of DHCP.

\begin{table}[htbp]
    \caption{Results of DHCP on Qwen2.5-VL-7B and out-of-distribution POPE-ImageNet.}
    \resizebox{\linewidth}{!}{
    \begin{tabular}{c|c|cccc}
        \toprule
        Training set & Metric & Precision & Recall & F1-score & Support \\
        \midrule
        \multirow{4.5}{*}{COCO} & Non-hallucination & 93.39 & 88.85 & 91.07 & 85749 \\
        \multirow{4}{*}{+ GQA}& Hallucination & 52.02 & 65.78 & 58.10 & 15753 \\
        \cmidrule(lr){2-6}
        \multirow{3.1}{*}{+ VG100K}& Macro avg & 72.71 & 77.32 & 74.58 & 101502 \\
        & Weighted avg & 86.97 & 85.27 & 85.95 & 101502 \\
        \cmidrule(lr){2-6}
        & Accuracy & \multicolumn{4}{c}{85.27} \\
        \midrule
        \multirow{5}{*}{COCO} & Non-hallucination & 92.76 & 85.99 & 89.25 & 85749 \\
        & Hallucination & 45.42 & 63.45 & 52.94 & 15753 \\
        \cmidrule(lr){2-6}
        & Macro avg & 69.09 & 74.72 & 71.09 & 101502 \\
        & Weighted avg & 85.41 & 82.49 & 83.61 & 101502 \\
        \cmidrule(lr){2-6}
        & Accuracy & \multicolumn{4}{c}{82.49} \\
        \bottomrule
        \end{tabular}
        
    }
    \label{tab:dhcp_pope_imagenet}
\end{table}

As for generative tasks, we utilized COCO-Caption as the training dataset for DHCP and evaluated its performance on Flickr30k. Since these two datasets are entirely non-overlapping, we can effectively evaluate DHCP's ability to detect generative hallucinations in out-of-distribution scenarios. We conducted experiments using Qwen2.5-VL-7B. \Cref{tab:dhcp_flickr30k} shows that DHCP demonstrates robust generalization performance, even when applied to out-of-distribution data and generative hallucinations.

\begin{table}[htbp]
    \caption{Results of DHCP on Qwen2.5-VL-7B and out-of-distribution Flickr30k trained on COCO-Caption.}
    \resizebox{\linewidth}{!}{
    \begin{tabular}{c|cccc}
        \toprule
        Metric & Precision & Recall & F1-score & Support \\
        \midrule
        Non-hallucination & 73.76 & 73.12 & 73.44 & 20295 \\
        Hallucination & 53.24 & 54.06 & 53.64 & 11488 \\
        \midrule
        Macro avg & 63.50 & 63.59 & 63.54 & 31783 \\
        Weighted avg & 66.34 & 66.23 & 66.29 & 31783 \\
        \midrule
        Accuracy & \multicolumn{4}{c}{66.23} \\
        \bottomrule
        \end{tabular}
    }
    \label{tab:dhcp_flickr30k}
\end{table}

\section{Metrics of DHCP}
To evaluate hallucination detection performance, we mainly use precision, recall, F1 scores, and accuracy. By treating hallucinatory samples as positive and non-hallucinatory samples as negative, we categorized predictions into true positives (TP), false positives (FP), true negatives (TN), and false negatives (FN) based on their actual and predicted labels. These metrics were computed using the equations in \cref{eq:accuracy,eq:precision,eq:f1scores}. Additionally, we reported macro-averaged and weighted-averaged metrics. The ``macro avg'' represents the arithmetic mean of performance across hallucination and non-hallucination classes, while the ``weighted avg'' provides a class-weighted average based on sample sizes.
Notably, our primary focus lies in detecting hallucinations, so we place greater emphasis on performance metrics for the hallucination category.
\begin{align}
    \text{Precision} &= \frac{\text{TP}}{\text{TP}+\text{FP}} \label{eq:precision}, \quad
    \text{Recall} = \frac{\text{TP}}{\text{TP}+\text{FN}},\\
    \text{Accuracy} &= \frac{\text{TP}+\text{TN}}{\text{TP}+\text{FP}+\text{TN}+\text{FN}} \label{eq:accuracy}, \\
    \text{F1} &= 2\times\frac{\text{Precision}\times \text{Recall}}{\text{Precision} + \text{Recall}}. \label{eq:f1scores}
\end{align}

\section{Examples of Instructions in AMBER}
\begin{table}[!htbp]
\centering
    \caption{Examples of instructions for evaluating various types of hallucinations in AMBER.}
    \label{tab:amber_instructions}
    \resizebox{\linewidth}{!}{
    \begin{tabular}{c|c}
        \toprule
        Hallucination type & Example instruction \\
        \midrule
        Object & Is there a streetlamp in this image? \\
        Relation & Is there direct contact between the woman and paddle? \\
        Attribute $\to$ state & Is the sky sunny in this image? \\
        Attribute $\to$ action & Does the woman laugh in this image? \\
        Attribute $\to$ number & Are there two men in this image? \\
        \bottomrule
        \end{tabular}
    }
\end{table}

\section{Visualization Presentation of DHCP}

\begin{figure*}[!htbp]
  \centering
    \begin{subfigure}{0.49\linewidth}
        \includegraphics[width=\linewidth]{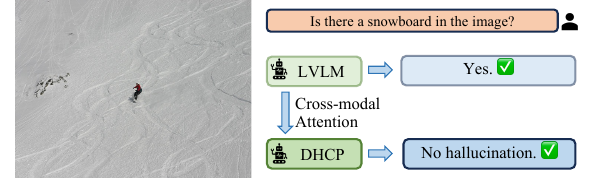}
        \caption{The non-hallucianation samples detected by DHCP on POPE-COCO.}
        \label{fig:coco_1}
    \end{subfigure}
    \hfill
    \begin{subfigure}{0.49\linewidth}
        \includegraphics[width=\linewidth]{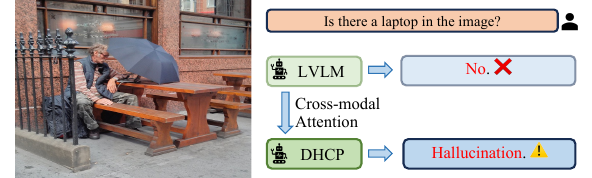}
        \caption{The hallucianation samples detected by DHCP on POPE-COCO.}
        \label{fig:coco_2}
    \end{subfigure}
    \hfill
    \begin{subfigure}{0.49\linewidth}
        \includegraphics[width=\linewidth]{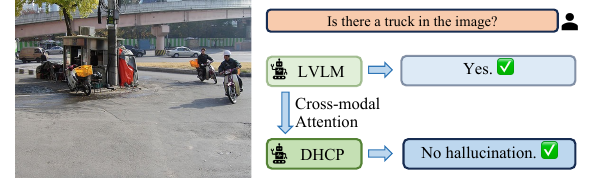}
        \caption{The non-hallucianation samples detected by DHCP on POPE-COCO.}
        \label{fig:coco_3}
    \end{subfigure}
    \hfill
    \begin{subfigure}{0.49\linewidth}
        \includegraphics[width=\linewidth]{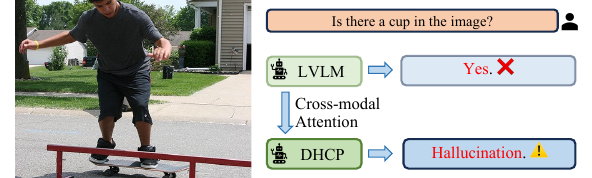}
        \caption{The hallucianation samples detected by DHCP on POPE-COCO.}
        \label{fig:coco_4}
    \end{subfigure}
    \hfill
    \begin{subfigure}{\linewidth}
        \includegraphics[width=\linewidth]{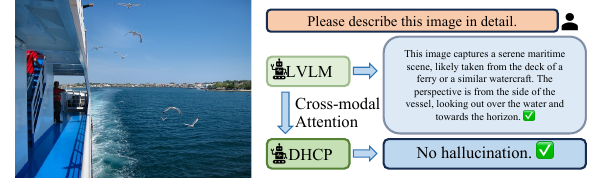}
        \caption{The non-hallucianation samples detected by DHCP on COCO-Caption.}
        \label{fig:caption_1}
    \end{subfigure}
    \hfill
    \begin{subfigure}{\linewidth}
        \includegraphics[width=\linewidth]{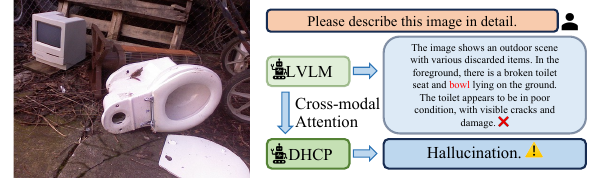}
        \caption{The hallucianation samples detected by DHCP on COCO-Caption.}
        \label{fig:caption_2}
    \end{subfigure}
  \caption {Visualization of DHCP for detecting hallucinations on POPE-COCO-test and COCO-Caption-test.}
  \label{fig:demo_coco}
\end{figure*}

\begin{figure*}[!htbp]
  \centering
    \begin{subfigure}{0.49\linewidth}
        \includegraphics[width=\linewidth]{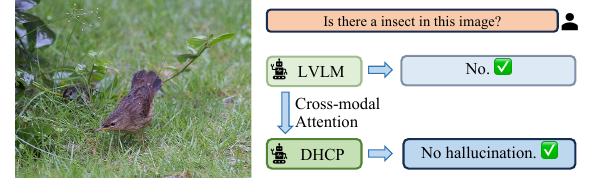}
        \caption{The non-hallucianation samples on AMBER-Object.}
        \label{fig:object_1}
    \end{subfigure}
    \hfill
    \begin{subfigure}{0.49\linewidth}
        \includegraphics[width=\linewidth]{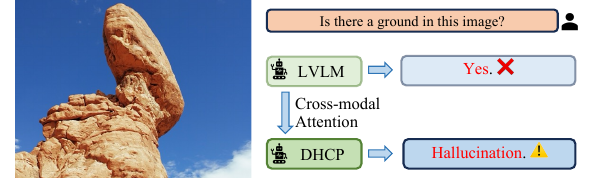}
        \caption{The non-hallucianation samples on AMBER-Object.}
        \label{fig:object_2}
    \end{subfigure}
    \hfill
    \begin{subfigure}{0.49\linewidth}
        \includegraphics[width=\linewidth]{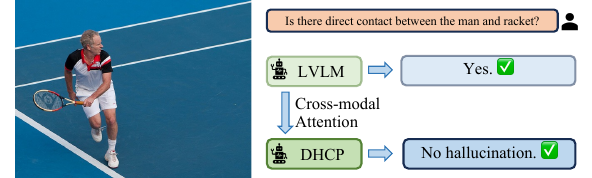}
        \caption{The hallucianation samples on AMBER-Relation.}
        \label{fig:relation_1}
    \end{subfigure}
    \hfill
    \begin{subfigure}{0.49\linewidth}
        \includegraphics[width=\linewidth]{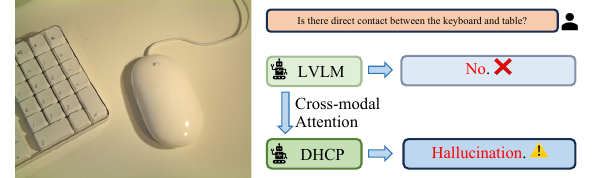}
        \caption{The non-hallucianation samples on AMBER-Relation.}
        \label{fig:relation_2}
    \end{subfigure}
    \hfill
    \begin{subfigure}{0.49\linewidth}
        \includegraphics[width=\linewidth]{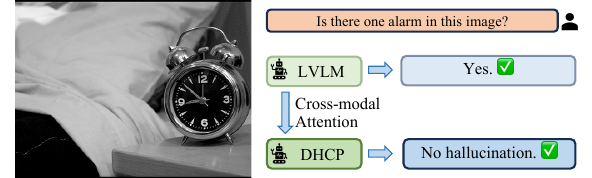}
        \caption{The non-hallucianation samples on AMBER-Attribute-Number.}
        \label{fig:number_1}
    \end{subfigure}
    \hfill
    \begin{subfigure}{0.49\linewidth}
        \includegraphics[width=\linewidth]{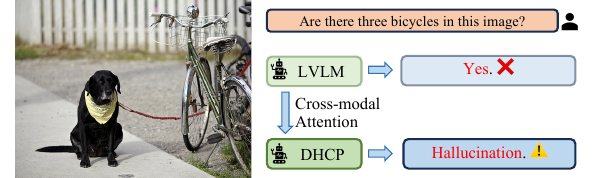}
        \caption{The hallucianation samples on AMBER-Attribute-Number.}
        \label{fig:number_2}
    \end{subfigure}
    \hfill
    \begin{subfigure}{0.49\linewidth}
        \includegraphics[width=\linewidth]{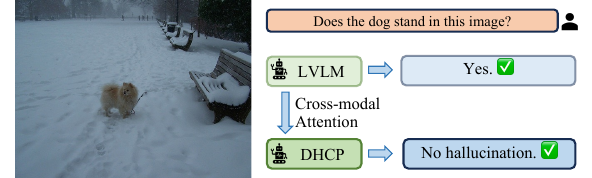}
        \caption{The non-hallucianation samples on AMBER-Attribute-Action.}
        \label{fig:action_1}
    \end{subfigure}
    \hfill
    \begin{subfigure}{0.49\linewidth}
        \includegraphics[width=\linewidth]{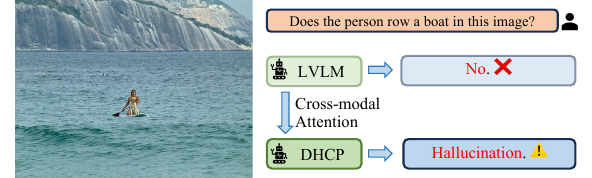}
        \caption{The hallucianation samples on AMBER-Attribute-Action.}
        \label{fig:action_2}
    \end{subfigure}
    \hfill
    \begin{subfigure}{0.49\linewidth}
        \includegraphics[width=\linewidth]{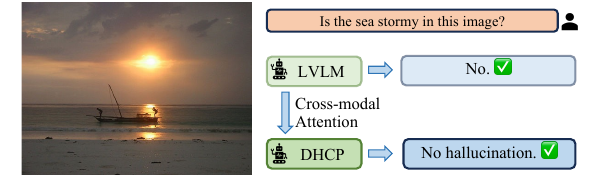}
        \caption{The non-hallucianation samples on AMBER-Attribute-State.}
        \label{fig:state_1}
    \end{subfigure}
    \hfill
    \begin{subfigure}{0.49\linewidth}
        \includegraphics[width=\linewidth]{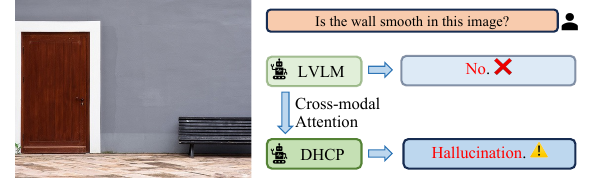}
        \caption{The hallucianation samples on AMBER-Attribute-State.}
        \label{fig:state_2}
    \end{subfigure}
    
  \caption {Visualization of DHCP for detecting hallucinations on AMBER-test.}
  \label{fig:demo_amber}
\end{figure*}

We show the visualization Presentation of DHCP in \cref{fig:demo_coco,fig:demo_amber} for POPE-COCO, COCO-caption and AMBER. 

\noindent
\textbf{POPE-COCO-test}. \Cref{fig:coco_1,fig:coco_2,fig:coco_3,fig:coco_4} illustrate the results of the hallucination detection on POPE-COCO-test.

\noindent
\textbf{COCO-Caption-test}. \Cref{fig:caption_1,fig:caption_2} illustrate the results of the hallucination detection on COCO-Caption-test.

\noindent
\textbf{AMBER-test}. \Cref{fig:demo_amber} illustrate the results of the hallucination detection on AMBER-test, including object hallucinations, relationship hallucinations and attribute hallucinations (state, number, action).

\end{document}